\documentclass{article}
\usepackage{spconf,amsmath,graphicx}
\usepackage{dsfont}
\usepackage{xcolor}
\usepackage{caption}
\usepackage{booktabs}
\usepackage{subcaption}
\usepackage{multirow}
\usepackage{multicol}

\title{Improving End-of-Turn Detection in Spoken Dialogues by Detecting 
Speaker Intentions as a Secondary Task}

\name{Zakaria Aldeneh$^1$, Dimitrios Dimitriadis$^{2*}$,
Emily Mower Provost$^1$}
\address{$^1$University of Michigan at Ann Arbor, $^2$Microsoft\\
\texttt{\{aldeneh,emilykmp\}@umich.edu},
\texttt{didimit@microsoft.com}}
\begin{document}
%
\maketitle
\begin{abstract}
This work focuses on the use of acoustic cues for 
modeling turn-taking in
dyadic spoken dialogues. Previous work has shown that 
speaker intentions
(e.g., asking a question, uttering a backchannel, etc.) 
can influence turn-taking behavior and are good 
predictors 
of turn-transitions in spoken dialogues.
However, speaker intentions are not readily available
for use by automated systems at run-time; making it difficult to use this 
information to anticipate a turn-transition.
To this end, we propose a multi-task neural approach for predicting 
turn-transitions and speaker intentions simultaneously.
Our results show that adding the auxiliary task of
speaker intention prediction improves the performance of 
turn-transition prediction in spoken dialogues, without 
relying on additional input features during run-time.
\end{abstract}

\makeatletter{\renewcommand*{\@makefnmark}{}
    \footnotetext{$^*$Work completed while at IBM T. J. Watson Research Center.}}

\section{Introduction}
\label{sec:intro}
Dialogue agents must be able to engage in human-like 
conversations in order to make interactions with spoken 
dialogue systems more natural and less rigid.
Turn-management is an essential component of 
conversations as it allows participants in a dialogue 
to exchange control of the floor.
Studies have shown that conversation partners rely on 
both syntactic and prosodic cues to anticipate 
turn-transitions
\cite{levinson2016turn,garrod2015use,gravano2011turn}.
Syntactic cues include keywords and semantics of an 
uttered sentence. 
Prosodic cues include the final intonation of a clause, 
pitch level, and speaking rate.
In this work, we assess the efficacy of using acoustic
cues for anticipating turn-switches in dyadic spoken 
dialogues. Given a single utterance, our goal is 
to use acoustic cues to predict if there will be a switch 
in speakers for the upcoming utterance or not.

Modern spoken dialogue systems generally rely on simple thresholding 
approaches for modeling turn-taking
\cite{maier2017towards,liu2017turn,parada2017improved}.   
However, turn-management is a complex phenomenon,
in which participants in a conversation rely on multiple cues to anticipate 
turn changes or end-of-turns. We anticipate that interactions between humans 
and machines can be improved if dialogue systems can accurately anticipate 
turn-switches in spoken conversations.

Turn-taking in conversations can take many forms. 
The two basic turn-taking functions 
are \textit{hold} and \textit{switch}. 
Given an utterance in a conversation, a hold indicates
that the next utterance will be uttered by the same 
speaker while a switch indicates that the next utterance
will be uttered by the other speaker in the conversation.
Turn-switches can be further divided into 
\textit{smooth} and \textit{overlapping} switches
\cite{gravano2011turn}.
Smooth switches occur when there is silence between
two consecutive utterances from two speakers. 
Overlapping switches occur when a speaker starts uttering 
a sentence before the other speaker finishes uttering 
his/her sentence.

Previous works built models that
used both acoustic and syntactic information to 
anticipate turn-changes to help make turn-management more
natural in spoken dialogue systems
\cite{gravano2011turn,liu2017turn}.
Gravano and Hirschberg showed that 
raising contours of intonation correlates with 
turn-transitions while 
flat intonations correlates with turn-holds
\cite{gravano2011turn}.
They also showed that certain keywords
(e.g.,~\textit{``you know\dots''}) and
textual completion have good correlations with
turn-management functions.
In addition to the usefulness of syntactic and acoustic 
cues for modeling turn-taking, 
previous work showed that speaker intentions
(e.g., ask a question, utter a backchannel, etc.) can 
be good predictors of turn-transitions in dialogues
\cite{meshorer2016using,guntakandla2015modelling}.
For example, a switch in speaker turns is more 
likely to occur after encountering a question than it is
to occur after encountering a statement.
Although speaker intentions (sometimes referred to as 
dialogue acts)
are useful for predicting turn-transitions
\cite{meshorer2016using,guntakandla2015modelling}, 
they require human annotations and are not readily 
available during run-time.

We propose the use of a multi-task Long Short-Term Memory 
(LSTM) network that takes in a sequence of acoustic 
frames from a given utterance and predicts
turn-transitions and speaker intentions simultaneously.
During training time, the network
is optimized with a joint loss function using 
ground-truth labels for turns and intentions. During
test time, the network makes two predictions, one of 
which can be discarded or used by other modules in
a spoken dialogue system.
The advantage is that this allows the 
model to use representations that encode information 
about speaker intentions for anticipating turns changes.
Our experiments demonstrate that adding the detection of
speaker intentions as a secondary task improves the 
performance of anticipating turn-transitions.

\section{Related Work}
\label{sec:related}
The problem of modeling turn-taking in conversations has 
been extensively studied in the literature. 
In this section, we give an overview of related works 
that focused on speech or textual interactions (i.e., no 
visual cues).
Our work complements previous work by showing that
a model that uses acoustic cues for predicting 
turn-switches benefits from
adding speaker intentions prediction as an auxiliary 
task using the multi-task learning framework.

One line of work looked at the use of acoustic and lexical
features for modeling turn-taking behavior 
\cite{maier2017towards,liu2017turn,masumura2017online,ishimoto2017end}. 
Liu et al.~\cite{liu2017turn}, Masumura et al.~\cite{masumura2017online}, 
and Ishimoto et al.~\cite{ishimoto2017end}
looked at the problem in Japanese conversations while 
Maier et al.~\cite{maier2017towards}
looked at the problem in German conversations.
Masumura et al.\ proposed 
using stacked time-asynchronous sequential networks for 
detecting end-of-turns given sequences of asynchronous 
features (e.g., MFCCs and words)~\cite{masumura2017online}.
Ishimoto et al.\ investigated the dependency between 
syntactic and prosodic features and showed that 
combining the two features is useful for predicting
end-of-turns~\cite{ishimoto2017end}.
Liu et al.\ built a Recurrent Neural Network (RNN) to 
classify a given utterance into four classes that relate 
to turn-taking behavior using joint acoustic and lexical
embeddings~\cite{liu2017turn}.
Finally, Maier et al.\ built an LSTM with a threshold-based decoding and 
studied the trade-off between latency and cut-in rate for end-of-turn 
detection in simulated real-time dialogues~\cite{maier2017towards}. 
The conclusion reached by this line of work was that 
end-of-turn detection models benefit from augmenting classifiers that use 
acoustic information with lexical information.

Another line of work focused solely on the acoustic 
modality, pointing out that using lexical features would
(1) require access to a speech recognition pipeline and 
(2) bias the classifiers due to varying prompt types
\cite{arsikere2015enhanced}.
Arsikere et al.\ compared the 
effectiveness of acoustic features 
(e.g., pitch trends, spectral constancy, etc.)
for predicting end-of-turns in two datasets that differed
in prompt type (one is slow and deliberate, the other is 
fast and spontaneous)~\cite{arsikere2015enhanced}.
They found that the same acoustic cues were useful for 
detecting end-of-turns for both prompt types.

A final line of work used dialogue act information 
when modeling turn-taking behavior
\cite{meshorer2016using, guntakandla2015modelling,heeman2017turn}.
Guntakandla and Nielsen built a 
turn-taking model that relied on transcribed segments, 
intention labels, speaker information, and change in 
speaker information to predict turn-transitions in 
dialogues~\cite{guntakandla2015modelling}.
Meshorer and Heeman used current and 
past speaker intention labels along with two new 
features, relative turn length and relative floor 
control, summarizing past speaker behavior for 
predicting turn-switches in dialogues~\cite{meshorer2016using}.
Finally, Heeman and Lunsford showed that 
turn-taking behavior not only depends on previous and 
upcoming speech act types, but also depends on 
the nature of a dialogue; suggesting that turn-taking 
events should be split into several groups depending on 
speech act types and the context of the dialogue
\cite{heeman2017turn}.

The works of Meshorer and Heeman~\cite{meshorer2016using}, 
Guntakandla and Nielsen~\cite{guntakandla2015modelling}, 
and Heeman and Lunsford~\cite{heeman2017turn}
suggested that speaker intentions can be useful for 
predicting turn-transitions.
However, speaker intentions are not readily obtainable 
from utterances and
require manual human annotations. 
We are interested in
studying how we can augment acoustic systems with speaker 
intention information, available during
training time, to improve performance of turn-transitions predictions.

\section{Problem Setup}
\label{sec:problem_setup}

We follow the work of Meshorer and Heeman~\cite{
meshorer2016using} and
represent 
a conversation between
two speakers as a sequence of utterances, taking the 
following form:
$$u_1, u_2, \ldots, u_N$$
where each $u_i$ is an utterance in the conversation. 
The sequence of utterances are sorted in terms of start 
talk time. 
Let $spkr(\cdot)$ be a 
function that returns the speaker of a given utterance. 
Given $u_i$, the goal is to predict whether the following 
statement is true or false:
$$spkr(u_i) \neq spkr(u_{i+1})$$
If the statement is true, then a turn-switch will take 
place and the other speaker will speak next. If the 
statement is false, then the current speaker will 
continue speaking.

Each utterance in the sequence represents a complete 
sentence, containing both acoustic and lexical cues, and
varies in duration. We assume that we know the end-points
of each utterance as in~\cite{meshorer2016using,guntakandla2015modelling,
arsikere2015enhanced}. Utterance end-points can be readily obtained from 
modern voice-activity detection algorithms or using an end-of-utterance 
detection systems (e.g.,~\cite{parada2017improved}). We leave the problem of
combining our multi-task model with end-of-utterance detection for future work 
and focus on the problem of predicting turn-switches from acoustic cues for a 
given utterance.

\section{Dataset and Features}
\label{sec:dataset_and_features}

\subsection{Dataset}
\label{sssec:dataset}
We use the Switchboard corpus~\cite{godfrey1992switchboard} to model
turn-taking behavior in spoken dialogues. The corpus 
consists of dyadic telephony conversations between 
participants who were asked to discuss various topics. 
We use the annotations provided by the Switchboard Dialog Act 
Corpus\footnote{https://github.com/cgpotts/swda} (SwDA),
since we are interested in utilizing speaker intentions 
(i.e., dialogue act types).
The goal of SwDA corpus was to extend the original 
Switchboard corpus with dialogue act types that summarize
turn information in the conversations. 

However, the SwDA corpus does not map dialogue acts to timing information 
in the original media files of
the Switchboard corpus. It only maps dialogue acts to
lexical and turn information. 
We augment the SwDA corpus with the NXT Switchboard corpus
\cite{calhoun2010nxt} to get utterance timing information
from the original media files.
The aim of the NXT Switchboard corpus was to combine major annotations that 
were performed on the Switchboard corpus and make them
accessible within one framework.

\textbf{Preparation}. 
We first add binary turn labels (switch/hold) 
to each utterance in the dataset.
We focus on $7$ major dialogue acts which we obtain by grouping
different SwDA classes as shown in Table~\ref{tab:intention_classes}. 
The dialogue act groups that we use are a subset of those
used in~\cite{meshorer2016using}.
We filtered out utterances in the dataset that do not
have corresponding audio segments (i.e., no timing
information).
We obtain the final utterances
by trimming the audio of the appropriate speaker channel
in the original media files in accordance to the timings
provided in the NXT Switchboard corpus.


\textbf{Analysis}. The final dataset that we use 
contains a total of 86,687 utterances.
Table~\ref{tab:turns_stats} shows a summary
of the content of the dataset in terms of turn labels and 
speaker intentions. As a first pass to understand the relationship
between dialogue acts and turns, we run a Chi-square test of independence 
and find that there is a relationship between dialogue acts and turns, 
$p<0.001$ (i.e., they are not independent). 
Note that this finding is suggestive rather than conclusive; mainly because
our utterances are not independent (they can come from same speaker).
Nevertheless, this finding supports findings in 
literature~\cite{gravano2011turn,heeman2017turn}, which suggested that speaker 
intentions influence turn-taking behavior.


\begin{table}[t]
\caption{Mapping dialogue act classes to intention classes.}
\label{tab:intention_classes}
\small
    \centering
    \begin{tabular}{ll}
        \toprule
        \textbf{SwDA classes} & \textbf{Intention classes} \\
        \midrule

        sd, h, bf & statement  \\
        sv, ad, sv@ & opinion  \\
        aa & agree \\
        \%, \%- & abandon \\
        b, bh & backchannel  \\
        qy, qo, qh & question \\
        no, ny, ng, arp & answer \\

        \bottomrule
    \end{tabular}
\end{table}

\begin{table}[t]
\caption{Total number of utterances that are followed by holds 
         and switches for each speaker intention class.}
\label{tab:turns_stats}
\small
    \centering
    \begin{tabular}{ccc}
        \toprule
        \multicolumn{1}{c}{\multirow{2}{*}{\textbf{Intention}}} & 
        \multicolumn{2}{c}{Counts ($\%$)} \\

        &
        \multicolumn{1}{c}{\textbf{Holds}} & 
        \multicolumn{1}{c}{\textbf{Switches}} \\

        \midrule

        statement   & 26,332 (52.2) & 12,722 (35.1) \\
        opinion     & 8,066 (16.0)  & 5,227 (14.4) \\
        agree       & 3,997 (7.9)   & 1,417 (3.9) \\
        abandon     & 3,887 (7.7)   & 3,203 (8.8) \\
        backchannel & 6,225 (12.3)  & 10,678 (29.5) \\
        question    & 752 (1.5)     & 2,369 (6.5) \\
        answer      & 1,197 (2.4)   & 615 (1.7) \\
        \bf{total}  & 50,456        & 36,231 \\

        \bottomrule
    \end{tabular}
\end{table}

\subsection{Features}
\label{sssec:features}
We use the OpenSMILE toolkit~\cite{eyben2010opensmile} to 
extract the following
features and their first (left) derivatives using a 
$25$ms Hamming window with a shift-rate of $10$ms:
intensity, loudness, MFCC, RMS energy, 
zero-crossing-rate, and smoothed pitch. 
As a result, a given signal is represented as a sequence 
of $42$-dimensional feature vectors.
The choice of these features was inspired by their 
success in previous studies on modeling turn-taking using 
acoustic cues~\cite{maier2017towards,arsikere2015enhanced}.

\section{Method}
\label{sec:method}
We use unidirectional LSTM network
to model the sequence of acoustic features and make
turn predictions. LSTMs are able to capture past signal
behavior and they have shown success in many audio
processing applications, such as speech recognition
and computational paralinguistics 
\cite{parada2017improved,brueckner2017spotting}.
In addition to their ability to capture past signal behavior,
LSTMs are able to capture information relating to timing 
and differentials (e.g., rising slope); both of which are useful for modeling 
turn-taking~\cite{gravano2011turn}.

Predicting turns can be formulated as a binary 
classification task where the goal is, given an utterance, 
predict whether there will be a 
turn-switch or a turn-hold. We augment this task given the 
following problem setup:
given an utterance, simultaneously predict turn-transitions and 
speaker intentions. 
The model is trained to minimize a
joint loss function that takes the following form:
$$L_{tot} = \lambda_1L_{turn}+\lambda_2L_{intent}$$
where $L_{turn}$ is the loss function for turn 
predictions, $L_{intent}$ is the loss function for
speaker intention predictions, $L_{tot}$ is the overall
loss function, $\lambda_1$ and $\lambda_2$ are weights
assigned to control the influence of each loss function.
In this work we set $\lambda_1$ to $1.0$ and $\lambda_2$ to $0.5$.

\textbf{Baselines.} We re-implement the ``full model'' from 
\cite{meshorer2016using} and compare its performance to 
the proposed
approach. The full model uses a Random Forest classifier 
with the features described in Section~\ref{sec:related}.
We also compare our proposed multi-task approach to a 
single-task LSTM that is trained to minimize $L_{turn}$ alone. 
We note that the single-task LSTM approach is similar to the one used 
in~\cite{maier2017towards}.

\section{Experiments}
\label{sec:experiments}

\subsection{Setup}
\label{sssec:setup}
We evaluate performance using 5-fold cross-validation.
We split on conversations, as opposed to utterances, to 
ensure that individual speakers do not appear in both the
training and testing folds.
For each testing fold, we randomly take out $33\%$ of the 
training conversations and use them for validation and 
early stopping.
For each conversation, we perform speaker-specific 
$z$-normalization on the features. 

We implement our models using the PyTorch 
library\footnote{https://github.com/pytorch/pytorch}.
We optimize the weighted negative log-likelihood loss 
function and use RMSProp optimizer to train our models. 
We use an initial learning rate of $0.001$.
At the end of each epoch, we compute the macro-F1 score 
on the validation set and reduce the learning rate by a 
factor of $2$ if there was no improvement
from last epoch. We run for a maximum of $100$ epochs
and stop training if there was no improvement in 
validation F1 score for 
$5$ consecutive epochs. We take a snapshot of the model 
after each epoch and select the one that gave the highest 
validation performance.

For each fold, we perform a grid search and pick the hyper-parameters that 
maximize validation performance. The main hyper-parameters of the model are:
number of layers $\{1, 2\}$ and layer width $\{32, 64, 128\}$. Once we  have
identified the optimal hyper-parameters for each fold, we train $3$ models 
with different random seeds and report their ensemble performance to minimize
variance due to random initialization. We report the average performance across
the five folds.

\subsection{Results and Discussion}
\label{sssec:results}

\begin{table}[t]
\caption{Performance comparison of different methods. 
Results shown are macro-averages across turn-switches and turn-holds.}
\label{tab:results_table}
\small
    \centering
    \begin{tabular}{lllll}
        \toprule
        \textbf{Method} & Rec. & Prec. & F1 & AUC \\
        \midrule

        Random & $50.0$ & $49.6$  & $41.7$  & $45.3$ \\
        Full model~\cite{meshorer2016using}  & 
                   $55.8$ & $56.7$ & $55.4$ & $57.8$ \\
        LSTM     & $65.9$ & $65.6$ & $65.5$ & $71.9$ \\
        MT-LSTM & $\textbf{66.4}^*$ & $\textbf{66.0}$ & $\textbf{65.8}$ & 
        $\textbf{72.6}^*$ \\

        \bottomrule
    \end{tabular}
    \\
    \footnotesize{$^*$ indicates $p<0.05$ under a paired $t$-test with LSTM.}
\end{table}

Table~\ref{tab:results_table} shows the results obtained 
from our experiments.
The table shows that a single-task LSTM, which uses the 
input features described in Section~\ref{sssec:features}, 
outperforms the 
full model in all evaluation metrics. We attribute this 
improvements to better feature representations and better 
sequential modeling abilities of LSTMs.
The table shows that a multi-task LSTM, which is trained
using a joint loss function, provides consistent 
improvements over a 
single-task LSTM\@~(significant improvements under a 
paired $t$-test, $p<0.05$, in terms of recall and AUC\@).
This suggests that a turn prediction model can 
benefit from representations extracted for detecting 
speaker intentions.
Next, we study how well our model is able to identify 
turn-switches when the switches are smooth and when they 
are overlapping. Our model identifies turn-switches with 
a recall of $68.5\%$ when the switches are overlapping 
and identifies turn-switches with a recall of $68.1\%$ 
when the switches are smooth.

Table~\ref{tab:intention_perf} shows the performance of 
predicting turn-switches and turn-holds for each 
intention class, as well as the accuracy of detecting that
intention class.
The results show that the model is better able to predict 
turn-switches when presented with a backchannel or a 
question, and is better able to predict turn-holds
when presented with a statement, opinion, or an answer.
This suggests that the performance of the model depends 
on the context and nature of a dialogue, and that it is easier to 
anticipate turn-switches or turn-holds for some intentions and not for others.

Table~\ref{tab:intention_perf} also shows the performance 
of identifying speaker intentions by the auxiliary 
task. The table shows that it is easier to identify
backchannels, questions, or turn-exit signals (abandon) than it is to 
identify agreement signals and statements.
The auxiliary task obtains an unweighted average recall 
(UAR) of $45.6\%$ on a $7$-way classification task 
(where chance UAR is $14.3\%$).

\begin{table}[t]
\caption{Detecting turn-switches and turn-holds for each speaker intention
class, as well as the per-class accuracy for detecting intentions by the 
auxiliary task.}
\label{tab:intention_perf}
\small
\centering
\begin{tabular}{cccc}
    \toprule
    \multicolumn{1}{c}{\multirow{2}{*}{\textbf{}}} & 
    \multicolumn{2}{c}{\textbf{Turn-Transitions}} & 
    \multicolumn{1}{c}{\textbf{Intentions}} \\
    \multicolumn{1}{c}{} & F1 (switch) & F1 (hold) &  per-class Acc.\\
    \midrule
        statement    &  $51.4$ & $73.2$ &  $39.6$ \\
        opinion      &  $54.5$ & $70.3$ &  $43.1$ \\
        agree        &  $49.7$ & $66.5$ &  $30.4$ \\
        abandon      &  $67.2$ & $68.3$ &  $56.8$ \\
        backchannel  &  $79.1$ & $51.6$ &  $49.6$ \\
        question     &  $72.2$ & $47.6$ &  $50.3$ \\
        answer       &  $61.1$ & $71.7$ &  $43.9$ \\
    \bottomrule
\end{tabular}
\end{table}

\section{Conclusion}
\label{sec:conclusion}
In this work we showed that a model that uses acoustic features for modeling 
turn-taking in spoken dialogues could benefit from adding speaker intention 
detection as an auxiliary task. We also explored how the performance of our 
turn-taking model varies depending on speaker intentions. For future work, 
we plan to augment acoustic features with lexical or phonetic information. 
We also plan to investigate combining turn-taking with end-of-utterance 
detection. Finally, we plan to add our model to a live spoken dialogue system.

\textbf{Acknowledgment.}
This work was supported by IBM under the Sapphire project. 


\bibliographystyle{IEEEbib}
\bibliography{strings,main}

\end{document}